\documentclass[letterpaper, conference]{IEEEtran}
\usepackage{times}

\usepackage{amsmath}
\usepackage{amsfonts}
\usepackage{amssymb}
\usepackage{balance}
\usepackage{bm}
\usepackage{booktabs}
\usepackage{hyperref}
\usepackage{xcolor}
\usepackage[super]{nth}
\usepackage{graphicx}
\usepackage{multicol}
\usepackage{multirow}
\usepackage[numbers, square, comma, sort&compress]{natbib}
\usepackage{pstool}
\usepackage[binary-units=true]{siunitx}
\usepackage[ruled,vlined]{algorithm2e}
\usepackage[normalem]{ulem}

\graphicspath{{./images/}}
\IEEEoverridecommandlockouts
\title{AlphaPilot: Autonomous Drone Racing}

\author{Philipp Foehn$^*$, Dario Brescianini$^*$, Elia Kaufmann$^*$,\\ Titus Cieslewski, Mathias Gehrig, Manasi Muglikar, Davide Scaramuzza
\thanks{* These authors contributed equally. All authors are with the Robotics and Perception Group, Dep. of Informatics, University of Zurich, Dep. of Neuroinformatics, University of Zurich and ETH Zurich, \url{http://rpg.ifi.uzh.ch/}
This work was supported by the Intel Network on Intelligent Systems, the Swiss National Science Foundation (SNSF) through the National Center of Competence in Research (NCCR) Robotics, the SNSF-ERC Starting Grant, and SONY.}}

\definecolor{somegray}{rgb}{0.5, 0.5, 0.5}
\newcommand{\darkgrayed}[1]{\textcolor{somegray}{#1}}
\makeatletter
\newcommand*\titleheader[1]{\gdef\@titleheader{#1}}
\AtBeginDocument{%
  \let\st@red@title\@title
  \def\@title{%
    \vskip-1.5em
    \bgroup\normalfont\large\centering\@titleheader\par\egroup
    \vskip0.5em\st@red@title}
}
\makeatother

\titleheader{\darkgrayed{This paper is an extended version of an accepted publication from Robotics: Science and Systems, 2020.\\
This version has been accepted for publication in Autonomous Robots (Springer).\\
Please cite as "AlphaPilot: Autonomous Drone Racing", P. Foehn, Autonomous Robots 2021.}}

\begin{document}
\newcommand{\myred}{red!15}
\newcommand{\mygreen}{green!15}
\newcommand{\myorange}{orange!15}
\newcommand{\myblue}{blue!15}

\newcommand{\I}{\mathcal{I}}    
\newcommand{\B}{\mathcal{B}}    
\newcommand{\G}{\mathcal{G}}    
\newcommand{\C}{\mathcal{C}}    
\renewcommand{\V}{\mathcal{V}}  

\newcommand{\pos}{\text{pos}}
\newcommand{\att}{\text{att}}
\newcommand{\cmd}{\text{cmd}}
\newcommand{\err}{\text{err}}
\newcommand{\des}{\text{ref}}
\newcommand{\ff}{\text{ff}}
\renewcommand{\i}{\text{i}}

\newcommand{\diff}[2]{\frac{\delta #1}{\delta #2}}
\newcommand{\part}[2]{\frac{\partial #1}{\partial #2}}
\newcommand{\mat}[1]{\begin{bmatrix} #1 \end{bmatrix}}
\renewcommand{\skew}[1]{\lbrack\bm{#1}\rbrack_{\!\times}}
\newcommand{\real}[0]{{\rm I\!R}}

\newcommand{\rom}[1]{\expandafter{\romannumeral #1\relax}}
\newcommand{\qmarks}[1]{``#1"}

\newcommand{\auro}[1]{#1}
\newcommand{\revision}[1]{#1}

\maketitle

\begin{abstract}
This paper presents a novel system for autonomous, vision-based drone racing combining learned data abstraction, nonlinear filtering, and time-optimal trajectory planning.
The system has successfully been deployed at the first autonomous drone racing world championship: the \textit{2019 AlphaPilot Challenge}.
Contrary to traditional drone racing systems, which only detect the next gate, our approach makes use of any visible gate  and takes advantage of multiple, simultaneous gate detections to compensate for drift in the state estimate and build a global map of the gates. The global map and drift-compensated state estimate allow the drone to navigate through the race course even when the gates are not immediately visible and further enable to plan a near time-optimal path through the race course in real time based on approximate drone dynamics.
The proposed system has been demonstrated to successfully guide the drone through tight race courses reaching speeds up to $\SI[detect-weight=true, per-mode=symbol]{8}{\meter\per\second}$ and ranked second at the \textit{2019 AlphaPilot Challenge}.
\end{abstract}

\IEEEpeerreviewmaketitle
~\\
\noindent
RSS paper:
{\footnotesize\url{http://rpg.ifi.uzh.ch/docs/RSS20_Foehn.pdf}}

\normalsize
\noindent
Video of our approach:
{\footnotesize\url{https://youtu.be/DGjwm5PZQT8}}

\noindent
Talk at RSS 2020:
{\footnotesize\url{https://youtu.be/k6vGEj1ZZWc}}

\section{Introduction}\label{sec:introduction}

\subsection{Motivation}
Autonomous drones have seen a massive gain in robustness in recent years and perform an increasingly large set of tasks across various commercial industries; however, they are still far from fully exploiting their physical capabilities.
Indeed, most autonomous drones only fly at low speeds near hover conditions in order to be able to robustly sense their environment and to have sufficient time to avoid obstacles.
Faster and more agile flight could not only increase the flight range of autonomous drones, but also improve their ability to avoid fast dynamic obstacles and enhance their maneuverability in confined spaces.
Human pilots have shown that drones are capable of flying through complex environments, such as race courses, at breathtaking speeds. However, autonomous drones are still far from human performance in terms of speed, versatility, and robustness, so that a lot of research and innovation is needed in order to fill this gap.

In order to push the capabilities and performance of autonomous drones,
in 2019, Lockheed Martin and the Drone Racing League have launched the \textit{AlphaPilot Challenge}\footnote{\url{https://thedroneracingleague.com/airr/}}$^{,}$\footnote{\url{https://www.nytimes.com/2019/03/26/technology/alphapilot-ai-drone-racing.html}}, an open innovation challenge with a grand prize of \$$1$~million.
The goal of the challenge is to develop a fully autonomous drone that navigates through a race course using machine vision, and which could one day beat the best human pilot.
While other autonomous drone races \cite{Moon16ram, Moon19springer} focus on complex navigation, the \textit{AlphaPilot Challenge} pushes the limits in terms of speed and course size to advance the state of the art and enter the domain of human performance.
Due to the high speeds at which drones must fly in order to beat the best human pilots, the challenging visual environments (e.g., low light, motion blur), and the limited computational power of drones, autonomous drone racing raises fundamental challenges in real-time state estimation, perception, planning, and control.

\begin{figure}[t]
	\centering
    \includegraphics[width=\linewidth,trim={0 40 0 10},clip]{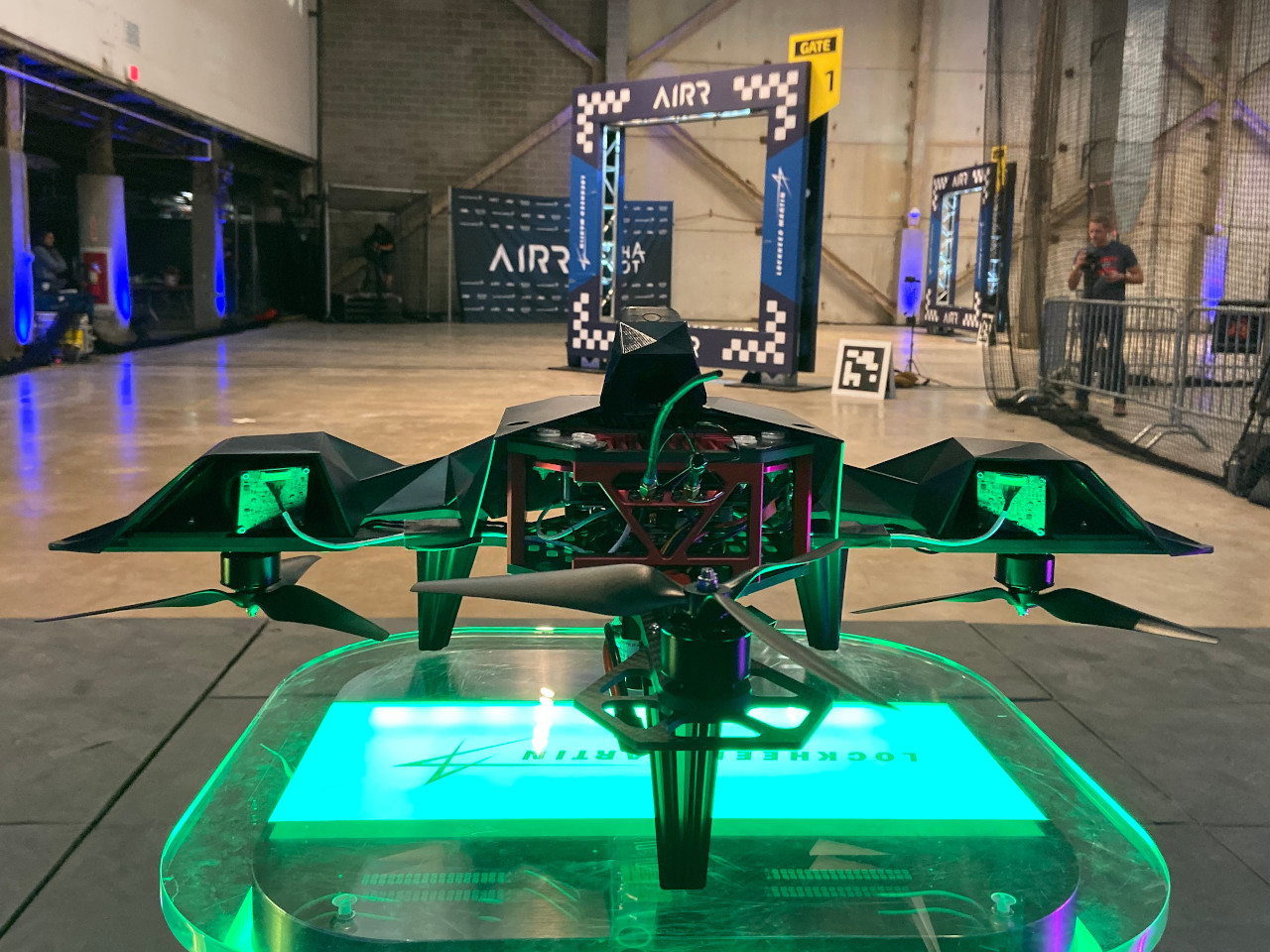}
	\caption{Our \textit{AlphaPilot} drone waiting on the start podium to autonomously race through the gates ahead.}
   \label{fig:drone_on_podium}
   \vspace{-0.35cm}
\end{figure}

\subsection{Related Work}
Autonomous navigation in indoor or GPS-denied environments typically relies on simultaneous localization and mapping (SLAM), often in the form of visual-inertial odometry (VIO)~\cite{Cadena16tro}.
There exists a variety of VIO algorithms, e.g., \cite{Mourikis07icra, Blosch15iros, Qin18tro, Forster17troSVO}, that are based on feature detection and tracking that achieve very good results in general navigation tasks~\cite{Delmerico18icra}.
However, the performance of these algorithms significantly degrades during agile and high-speed flight as encountered in drone racing.
The drone's high translational and rotational velocities cause large optic flow, making robust feature detection and tracking over sequential images difficult and thus causing substantial drift in the VIO state estimate~\cite{Delmerico19icra}.

To overcome this difficulty, several approaches exploiting the structure of drone racing with gates as landmarks have been developed, e.g.,~\cite{Li19arxiv,Jun18ral, Kaufmann18icra}, where the drone locates itself relative to gates.
In~\cite{Li19arxiv}, a handcrafted process is used to extract gate information from images that is then fused with attitude estimates from an inertial measurement unit (IMU) to compute an attitude reference that guides the drone towards the visible gate.
While the approach is computationally very light-weight, it struggles with scenarios where multiple gates are visible and does not allow to employ more sophisticated planning and control algorithms which, e.g., plan several gates ahead.
In \cite{Jun18ral}, a convolutional neural network (CNN) is used to retrieve a bounding box of the gate and a line-of-sight-based control law aided by optic flow is then used to steer the drone towards the detected gate.
\auro{While this approach is successfully deployed on a real robotic system, the generated control commands do not account for the underactuated system dynamics of the quadrotor, constraining this method to low-speed flight.}
The approach presented in~\cite{Kaufmann18icra} also relies on relative gate data but has the advantage that it works even when no gate is visible. In particular, it uses a CNN to directly infer relative gate poses from images and fuse the results with a VIO state estimate. However, the CNN does not perform well when multiple gates are visible as it is frequently the case for drone racing.

\auro{
Assuming knowledge of the platform state and the environment, 
there exist many approaches which can reliably generate feasible trajectories with high efficiency.
The most prominent line of work exploits the quadrotor's underactuated nature and the resulting differentially-flat output states \cite{Mellinger12ijrr, Mueller15tro}, where trajectories are described as polynomials in time.
Other approaches additionally incorporate obstacle avoidance~\cite{Zhou19ral, Gao19jfr} or perception constraints~\cite{Falanga18iros, Spasojevic20icra}.
However, in the context of drone racing, specifically the AlphaPilot Challenge, obstacle avoidance is often not needed, but time-optimal planning is of interest.
There exists a handful of approaches for time-optimal planning~\cite{Hehn12ar, Loock13ecc, Ryou2020rss, Foehn2020cpc}.
However, while~\cite{Hehn12ar, Loock13ecc} are limited to 2D scenarios and only find trajectories between two given states, \cite{Ryou2020rss} requires simulation and real-world data obtained on the track, and the method of \cite{Foehn2020cpc} is not applicable due to computational constraints.
}

\subsection{Contribution}
The approach contributed herein builds upon the work of~\cite{Kaufmann18icra} and fuses VIO with a robust CNN-based gate corner detection using an extended Kalman filter (EKF), achieving high accuracy at little computational cost.
The gate corner detections are used as static features to compensate for the VIO drift and to align the drone's flight path precisely with the gates.
Contrary to all previous works \cite{Li19arxiv, Jun18ral, Kaufmann18icra}, which only detect the next gate, our approach makes use of any gate detection and even profits from multiple simultaneous detections to compensate for VIO drift and build a global gate map.
The global map allows the drone to navigate through the race course even when the gates are not immediately visible and further enables the usage of sophisticated path planning and control algorithms.
In particular, a computationally efficient, sampling-based path planner (see e.g.,~\cite{lavalle2006planning}, and references therein) is employed that plans near time-optimal paths through multiple gates ahead and is capable of adjusting the path in real time if the global map is updated.

\auro{
This paper extends our previous work \cite{Foehn20rss} by including a more detailed elaboration on our gate corner detection in Sec.~\ref{sec:gate_detection} with an ablation study in Sec.~\ref{sec:res_gatedetection}, further details on the fusion of VIO and gate detection in Sec.~\ref{sec:state_estimation}, and a description of the path parameterization in Sec.~\ref{sec:path_planning}, completed by an ablation study on the planning horizon length in Sec.~\ref{sec:res_planning}.
}
\section{AlphaPilot Race Format and Drone}\label{sec:description}

\begin{figure}[t]
        \centering
        \def\svgwidth{0.9\columnwidth}
        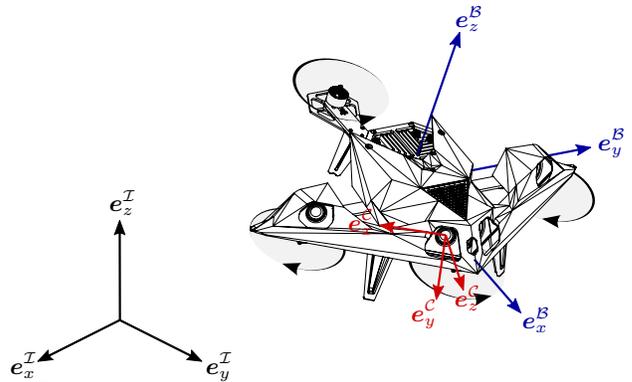
        \vspace{-0.35cm}
        \caption{Illustration of the race drone with its body-fixed coordinate frame $\mathcal{B}$ in blue and a camera coordinate frame $\mathcal{C}$ in red.}
        \label{fig:drone_dynamics}
        \vspace{-0.35cm}
\end{figure}

\begin{table}[b]
    \caption{Sensor specifications.}
    \label{tab:sensor_specifications}
    \small
    \centering
    \setlength{\tabcolsep}{3pt}
    \begin{tabular}{l l l l}
        \toprule
        Sensor & Model & Rate & Details \\
        \midrule
        Cam &
        \begin{tabular}{@{}l@{}} Leopard Imaging \\ IMX 264 \end{tabular}&
        $\SI{60}{\hertz}$ &
        \begin{tabular}{@{}l@{}} global shutter, color \\ resolution: $1200\times720$ \end{tabular} \\
        \midrule
        IMU &
        Bosch BMI088 &
        $\SI{430}{\hertz}$ &
        \begin{tabular}{@{}l@{}} range: $\pm 24 g$, $\pm\SI[per-mode=symbol]{34.5}{\radian\per\second}$ \\ resolution: $7\mathrm{e}^{\text{-}4}g$, $1\mathrm{e}^{\text{-}3}\si[per-mode=symbol]{\radian\per\second}$ \end{tabular} \\
        \midrule
        LRF &
        \begin{tabular}{@{}l@{}} Garmin \\ {LIDAR}-Lite v3 \end{tabular} &
        $\SI{120}{\hertz}$ &
        \begin{tabular}{@{}l@{}} range: $1$-$\SI{40}{\meter}$ \\ resolution: $\SI{0.01}{\meter}$ \end{tabular}\\
        \bottomrule
    \end{tabular}
\end{table}

\begin{figure*}[ht!]
        \centering
        \def\svgwidth{0.95\textwidth}
        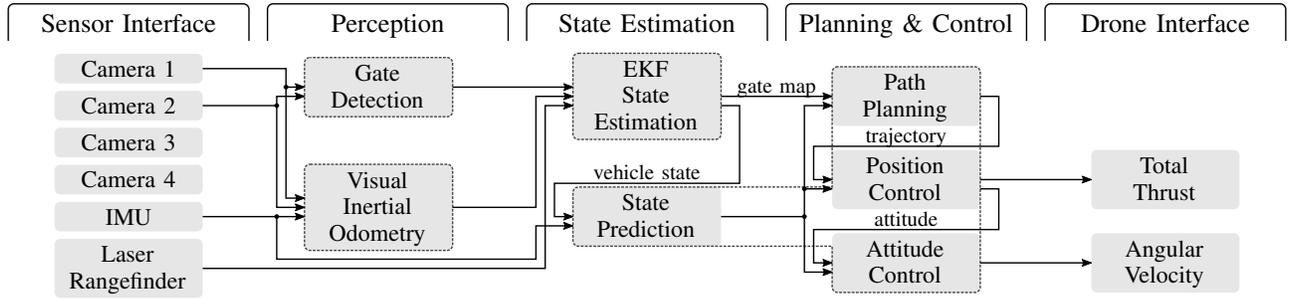
        \caption{Overview of the system architecture and its main components. All components within a dotted area run in a single thread.}
        \label{fig:system_overview}
        \vspace{-0.35cm}
\end{figure*}

\subsection{Race Format}
From more than 400 teams that participated in a series of qualification tests including a simulated drone race~\cite{guerra2019flightgoggles}, the top nine teams were selected to compete in the \textit{2019 AlphaPilot Challenge}.
The challenge consists of three qualification races and a final championship race at which the six best teams from the qualification races compete for the grand prize of \$$1$~million.
Each race is implemented as a time trial competition in which each team is given three attempts to fly through a race course as fast a possible without competing drones on the course.
Taking off from a start podium, the drones have to autonomously navigate through a sequence of gates with distinct appearances in the correct order and terminate at a designated finish gate.
The race course layout, gate sequence, and position are provided ahead of each race up to approximately $\pm\SI{3}{\meter}$ horizontal uncertainty, enforcing teams to come up with solutions that adapt to the real gate positions.
Initially, the race courses were planned to have a lap length of approximately $\SI{300}{\meter}$ and required the completion up to three laps.
However, due to technical difficulties, no race required to complete multiple laps and the track length at the final championship race was limited to about $\SI{74}{\meter}$.

\subsection{Drone Specifications}\label{sec:drones_specifications}
All teams were provided with an identical race drone (Fig.~\ref{fig:drone_on_podium}) that was approximately $\SI{0.7}{\meter}$ in diameter, weighed $\SI{3.4}{\kilo\gram}$, and had a thrust-to-weight ratio of $1.4$.
The drone was equipped with a NVIDIA Jetson Xavier embedded computer for interfacing all sensors and actuators and handling all computation for autonomous navigation onboard.
The sensor suite included two $\pm$\ang{30} forward-facing stereo camera pairs (Fig. \ref{fig:drone_dynamics}), an IMU, and a downward-facing laser rangefinder (LRF).
All sensor data were globally time stamped by software upon reception at the onboard computer.
Detailed specifications of the available sensors are given in Table~\ref{tab:sensor_specifications}.
The drone was equipped with a flight controller that controlled the total thrust $f$ along the drone's $z$-axis (see Fig.~\ref{fig:drone_dynamics}) and the angular velocity, $\bm{\omega}=\left(\omega_x,\omega_y,\omega_z\right)$, in the body-fixed coordinate frame $\B$.

\subsection{Drone Model}\label{sec:dronemodel}
Bold lower case and upper case letters will be used to denote vectors and matrices, respectively. The subscripts in ${_\I\bm{p}_{CB} = {_\I\bm{p}}_{B} - {_\I\bm{p}}_{C}}$ are used to express a vector from point $C$ to point $B$ expressed in frame $\I$. Without loss of generality, $I$ is used to represent the origin of frame $\I$, and $B$ represents the origin of coordinate frame $\B$. For the sake of readability, the leading subscript may be omitted if the frame in which the vector is expressed is clear from context.

The drone is modelled as a rigid body of mass $m$ with rotor drag proportional to its velocity acting on it~\cite{kai2017drag}.
The translational degrees-of-freedom are described by the position of its center of mass, $\bm{p}_{B}=\left(p_{B\!,x}, p_{B\!,y}, p_{B\!,z}\right)$, with respect to an inertial frame $\I$ as illustrated in Fig.~\ref{fig:drone_dynamics}. 
The rotational degrees-of-freedom are parametrized using a unit quaternion, $\bm{q}_{\I\B}$, where $\bm{R}_{\I\B}=\bm{R}(\bm{q}_{\I\B})$ denotes the rotation matrix mapping a vector from the body-fixed coordinate frame $\B$ to the inertial frame $\I$~\cite{shuster1993attitude}.
A unit quaternion, $\bm{q}$, consists of a scalar $q_w$ and a vector $\tilde{\bm{q}}=\left(q_x, q_y, q_z\right)$ and is defined as $\bm{q}=\left(q_w, \tilde{\bm{q}}\right)$~\cite{shuster1993attitude}.
The drone's equations of motion are
\begin{align}
m\ddot{\bm{p}}_{B} &= \bm{R}_{\I\B} f \bm{e}_z^\B - \bm{R}_{\I\B}\bm{D}\bm{R}_{\I\B}^\intercal \bm{v}_{B} -m\bm{g},\label{eq:trans_dynamics}\\
\dot{\bm{q}}_{\I\B} &= \frac{1}{2} \mat{0 \\ \bm{\omega}} \otimes \bm{q}_{\I\B},\label{eq:att_kinematics}
\end{align}
where $f$ and $\bm{\omega}$ are the force and bodyrate inputs, $\bm{e}_z^\B=\left(0,0,1\right)$ is the drone's $z$-axis expressed in its body-fixed frame $\B$, ${\bm{D} = \mathrm{diag}(d_x, d_y, 0)}$ is a constant diagonal matrix containing the rotor drag coefficients, $\bm{v}_B = \dot{\bm{p}}_B$ denotes the drone's velocity, $\bm{g}$ is gravity and $\otimes$ denotes the quaternion multiplication operator~\cite{shuster1993attitude}.
The drag coefficients were identified experimentally to be $d_x = \SI[per-mode=symbol]{0.5}{\kilo\gram\per\second}$ and $d_y = \SI[per-mode=symbol]{0.25}{\kilo\gram\per\second}$.

\section{System Overview}\label{sec:sys_overview}
The system is composed of five functional groups: Sensor interface, perception, state estimation, planning and control, and drone interface (see Fig.~\ref{fig:system_overview}).
In the following, a brief introduction to the functionality of our proposed perception, state estimation, and planning and control system is given.

\subsection{Perception}
Of the two stereo camera pairs available on the drone, only the two central forward-facing cameras are used for gate detection (see Section~\ref{sec:gate_detection}) and, in combination with IMU measurements, to run VIO. The advantage is that the amount of image data to be processed is reduced while maintaining a very large field of view.
Due to its robustness, multi-camera capability and computational efficiency, ROVIO~\cite{Blosch15iros} has been chosen as VIO pipeline. At low speeds, ROVIO is able to provide an accurate estimate of the quadrotor vehicle's pose and velocity relative to its starting position, however, at larger speeds the state estimate suffers from drift.

\subsection{State Estimation}
In order to compensate for a drifting VIO estimate, the output of the gate detection and VIO are fused together with the measurements from the downward-facing laser rangefinder (LRF) using an EKF (see Section~\ref{sec:state_estimation}).
The EKF estimates a global map of the gates and, since the gates are stationary, uses the gate detections to align the VIO estimate with the global gate map, i.e., compensates for the VIO drift.
Computing the state estimate, in particular interfacing the cameras and running VIO, introduces latency in the order of $\SI{130}{\milli\second}$ to the system. In order to be able to achieve a high bandwidth of the control system despite large latencies, the vehicle's state estimate is predicted forward to the vehicle's current time using the IMU measurements.

\subsection{Planning and Control}
The global gate map and the latency-compensated state estimate of the vehicle are used to plan a near time-optimal path through the next $N$ gates starting from the vehicle's current state (see Section~\ref{sec:path_planning}).
The path is re-planned every time (\rom{1}) the vehicle passes through a gate, (\rom{2}) the estimate of the gate map or (\rom{3}) the VIO drift are updated significantly, i.e., large changes in the gate positions or VIO drift.
The path is tracked using a cascaded control scheme (see Section~\ref{sec:control}) with an outer proportional-derivative (PD) position control loop and an inner proportional (P) attitude control loop.
Finally, the outputs of the control loops, i.e., a total thrust and angular velocity command, are sent to the drone.

\subsection{Software Architecture}
The NVIDIA Jetson Xavier provides eight CPU cores, however, four cores are used to run the sensor and drone interface.
The other four cores are used to run the gate detection, VIO, EKF state estimation, and planning and control, each in a separate thread on a separate core.
All threads are implemented asynchronously to run at their own speed, i.e., whenever new data is available, in order to maximize data throughput and to reduce processing latency.
The gate detection thread is able to process all camera images in real time at \SI{60}{\hertz}, whereas the VIO thread only achieves approximately \SI{35}{\hertz}.
In order to deal with the asynchronous nature of the gate detection and VIO thread and their output, all data is globally time stamped and integrated in the EKF accordingly.
The EKF thread runs every time a new gate or LRF measurement is available.
The planning and control thread runs at a fixed rate of \SI{50}{\hertz}. To achieve this, the planning and control thread includes the state prediction which compensates for latencies introduced by the VIO.

\section{Gate Detection}\label{sec:gate_detection}

\begin{figure*}[t]
    \centering
    \includegraphics[width=1.0\linewidth]{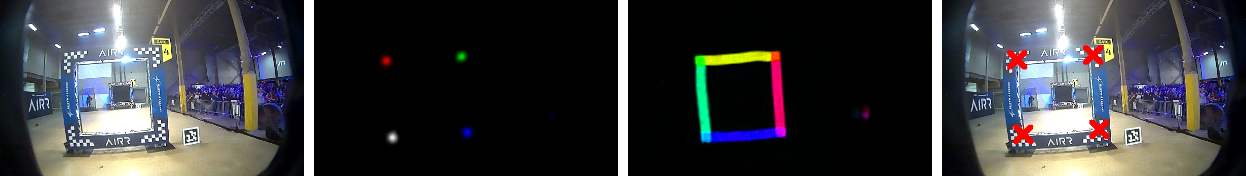}
    \par\medskip 
    \includegraphics[width=1.0\linewidth]{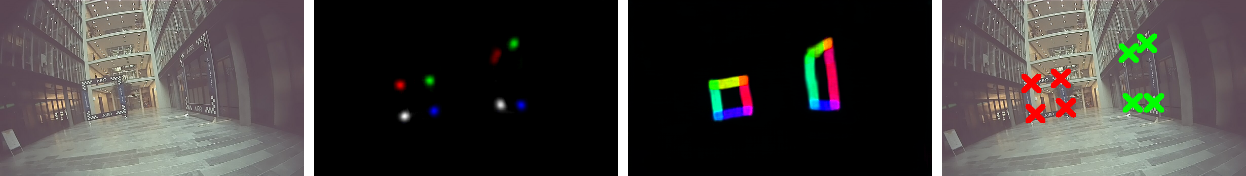}
    \par\medskip
    \includegraphics[width=1.0\linewidth]{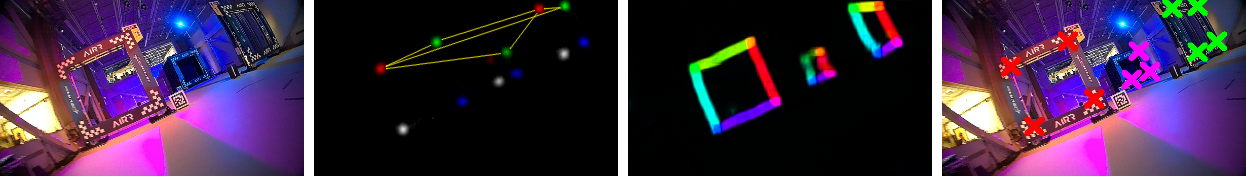}
    \caption{
    The gate detection module returns sets of corner points for each gate in the input image (fourth column) using a two-stage process.
    In the first stage, a neural network transforms an input image, $\bm{I}_{w\times h\times 3}$ (first column), into a set of confidence maps for corners, $\bm{C}_{w\times h\times 4}$ (second column), and Part Affinity Fields (PAFs)~\cite{cao2017realtime}, $\bm{E}_{w\times h\times (4 \cdot 2)}$ (third column).
    In the second stage, the PAFs are used to associate sets of corner points that belong to the same gate.
    For visualization, both corner maps, $\bm{C}$ (second column), and PAFs, $\bm{E}$ (third column), are displayed in a single image each.
    While color encodes the corner class for $\bm{C}$, it encodes the direction of the 2D vector fields for $\bm{E}$. 
    The yellow lines in the bottom of the second column show the six edge candidates of the edge class $({\color{red} TL}, {\color{green} TR})$ (the $TL$ corner of the middle gate is below the detection threshold), see Section~\ref{pafuse}.
    Best viewed in color.
    }
    \label{fig:network_prediction}
    \vspace{-0.35cm}
\end{figure*}

To correct for drift accumulated by the VIO pipeline, the gates are used as distinct landmarks for relative localization.
In contrast to previous CNN-based approaches to gate detection, we do not infer the relative pose to a gate directly, but instead segment the four corners of the observed gate in the input image.
\revision{
These corner segmentations represent the likelihood of a specific gate corner to be present at a specific pixel coordinate. 
To represent a value proportional to the likelihood, the maps are trained on Gaussians of the corner projections. 
}
This allows the detection of an arbitrary amount of gates, and allows for a more principled inclusion of gate measurements in the EKF through the use of reprojection error.
\revision{Specifically, it exhibits more predictable behavior for partial gate observations and overlapping gates, and allows to suppress the impact of Gaussian noise by having multiple measurements relating to the same quantity.}
Since the exact shape of the gates is known, detecting a set of characteristic points per gate allows to constrain the relative pose.
For the quadratic gates of the \textit{AlphaPilot Challenge}, these characteristic points are chosen to be the inner corner of the gate border (see Fig.~\ref{fig:network_prediction}, 4th column).
However, just detecting the four corners of all gates is not enough.
If just four corners of several gates are extracted, the association of corners to gates is undefined (see Fig.~\ref{fig:network_prediction}, 3rd row, 2nd column).
To solve this problem, we additionally train our network to extract so-called Part Affinity Fields (PAFs), as proposed by~\cite{cao2017realtime}.
These are vector fields, which, in our case, are defined along the edges of the gates, and point from one corner to the next corner of the same gate, see column three in Figure~\ref{fig:network_prediction}.
\auro{The entire gate detection pipeline consists of two stages: 1)~predicting corner maps and PAFs by the neural network, 2)~extracting single edge candidates from the network prediction and assembling them to gates. 
In the following, both stages are explained in detail.
}

\subsection{Stage 1: Predicting Corner Maps and Part Affinity Fields}
In the first detection stage, each input image, $\bm{I}_{w\times h\times 3}$,
is mapped by a neural network into a set of $N_C=4$ corner maps, $\bm{C}_{w\times h\times N_C}$,
and $N_E=4$ PAFs, $\bm{E}_{w\times h\times (N_E\cdot 2)}$.
\auro{Predicted corner maps as well as PAFs are illustrated in Figure~\ref{fig:network_prediction}, 2nd and 3rd column. 
}
The network is trained in a supervised manner by minimizing the Mean-Squared-Error loss between the network prediction and the ground-truth maps.
In the following, ground-truth maps for both map types are explained in detail.

\subsubsection{Corner Maps}
For each corner class, ${j \in \mathcal{C}_j}$, ${\mathcal{C}_j := \lbrace _{TL}, _{TR}, _{BL}, _{BR} \rbrace}$, a ground-truth corner map, $\bm{C}^\ast_{j}(\bm{s})$, is represented by a single-channel map of the same size as the input image and indicates the existence of a corner of class $j$ at pixel location $\bm{s}$ in the image.
The value at location $\bm{s} \in \bm{I}$ in $\bm{C}_{j}^\ast$ is defined by a Gaussian as
\begin{equation}
   \bm{C}^\ast_{j}(\bm{s}) = \exp \left( - \frac{\Vert \bm{s} - \bm{s}^\ast_{j} \Vert_2^2}{\sigma^2} \right), 
\end{equation}
where $\bm{s}^\ast_{j}$ denotes the ground truth image position of the nearest corner with class $j$. 
The choice of the parameter $\sigma$ controls the width of the Gaussian.
We use $\sigma = 7$ pixel in our implementation.
Gaussians are used to account for small errors in the ground-truth corner positions that are provided by hand.
\auro{Ground-truth corner maps are generated for each individual gate in the image separately and then aggregated. 
Aggregation is performed by taking the pixel-wise maximum of the individual corner maps, as this preserves the distinction between close corners.}

\subsubsection{Part Affinity Fields}
We define a PAF for each of the four possible classes of edges, defined by its two connecting corners as $(k, l) \in \mathcal{E}_{KL} := \lbrace (_{TL}, _{TR}), (_{TR}, _{BR}), (_{BR}, _{BL}), (_{BL}, _{TL}) \rbrace$.
For each edge class, $(k, l)$, the ground-truth PAF, $\bm{E}^\ast_{(k, l)}(\bm{s})$, is represented by a two-channel map of the same size as the input image and points from corner $k$ to corner $l$ of the same gate, provided that the given image point $\bm{s}$ lies within distance $d$ of such an edge.
We use $d = 10$ pixel in our implementation.
Let $\mathcal{G}^\ast$ be the set of gates $g$ and $\bm{S}_{(k, l), g}$ be the set of image points that are within distance $d$ of the line connecting the corner points $\bm{s}^\ast_{k}$ and $\bm{s}^\ast_{l}$ belonging to gate $g$.
Furthermore, let $\bm{v}_{k, l, g}$ be the unit vector pointing from $\bm{s}^\ast_{k}$ to $\bm{s}^\ast_{l}$ of the same gate.
Then, the part affinity field, $\bm{E}^\ast_{(k, l)}(\bm{s})$, is defined as:
\begin{equation}
\bm{E}^\ast_{(k, l)}(\bm{s}) = 
\begin{cases}
   \bm{v}_{k, l, g} \quad \text{if $\bm{s}\in \bm{S}_{(k, l), g}, \quad g \in \mathcal{G}^\ast$} \\
   \bm{0} \quad \text{otherwise.}
\end{cases}
\end{equation}
\auro{As in the case of corner maps, PAFs are generated for each individual gate in the image separately and then aggregated. 
In case a point $\bm{s}$ lies in $\bm{S}_{(k, l), g}$ of several gates, the $\bm{v}_{k, l, g}$ of all corresponding gates are averaged}.

\subsection{Stage 2: Corner Association}\label{pafuse}
At test time, discrete corner candidates, $\bm{s}_{j}$, for each corner class, $j \in \mathcal{C}_j$, are extracted from the predicted corner map using non-maximum suppression and thresholding.
\auro{For each corner class, there might be several corner candidates, due to multiple gates in the image or false positives.}
These corner candidates allow the formation of an exhaustive set of edge candidates, $\lbrace (\bm{s}_{k}, \bm{s}_{l}) \rbrace$, see the yellow lines in Fig.~\ref{fig:network_prediction}.
Given the corresponding PAF, $\bm{E}_{(k, l)}(\bm{s})$, each edge candidate is assigned a score which expresses the agreement of that candidate with the PAF.
This score is given by the line integral
\begin{equation}\label{eq:line_integral}
 \mathcal{S}((\bm{s}_{k}, \bm{s}_{l})) = \int_{u=0}^{u=1} \bm{E}_{(k, l)}(\bm{s}(u)) \cdot \frac{\bm{s}_{l}- \bm{s}_{k}}{\Vert\bm{s}_{l}- \bm{s}_{k}\Vert} du, 
\end{equation}
where $\bm{s}(u)$ lineraly interpolates between the two corner candidate locations $\bm{s}_k$ and $\bm{s}_l$.
In practice, $\mathcal{S}$ is approximated by uniformly sampling the integrand.

\auro{
The line integral $\mathcal{S}$ is used as metric to associate corner candidates to gate detections.
The goal is to find the optimal assignment for the set of all possible corner candidates to gates.
As described in~\cite{cao2017realtime}, finding this optimal assignment corresponds to a $K$-dimensional matching problem that is known to be NP-Hard~\cite{west2001introduction}. 
Following~\cite{cao2017realtime}, the problem is simplified by decomposing the matching problem into a set of bipartite matching subproblems. 
Matching is therefore performed independently for each edge class.
Specifically, the following optimization problem represents the bipartite matching subproblem for edge class $(k, l)$:
\begin{align}
    \max \mathcal{S}_{(k,l)} &= \sum_{k\in \mathcal{D}_k} \sum_{l\in \mathcal{D}_l}   \mathcal{S}((\bm{s}_{k}, \bm{s}_{l})) \cdot z_{kl} \\
    \text{s.t.} \quad &\forall k \in \mathcal{D}_k, \; \sum_{l\in \mathcal{D}_l} z_{kl} \leq 1 \; , \label{eq:match_constraint_1}\\
    &\forall l \in \mathcal{D}_l, \; \sum_{k\in \mathcal{D}_k} z_{kl} \leq 1 \; , \label{eq:match_constraint_2}
\end{align}
where $\mathcal{S}_{(k,l)}$ is the cumulative matching score and $\mathcal{D}_k$, $\mathcal{D}_l$ denote the set of corner candidates for edge class~$(k,l)$. 
The variable $z_{kl} \in \lbrace 0, 1\rbrace$ indicates whether two corner candidates are connected. 
Equations~\eqref{eq:match_constraint_1} and \eqref{eq:match_constraint_2} enforce that no two edges share the same corner. 
Above optimization problem can be solved using the Hungarian method~\cite{kuhn1955hungarian}, resulting in a set of edge candidates for each edge class $(k,l)$. 

With the bipartite matching problems being solved for all edge classes,
the pairwise associations can be extended to sets of associated edges for each gate.
}

\auro{
\subsection{Training Data}
The neural network is trained in a supervised fashion using a dataset recorded in the real world.
Training data is generated by recording video sequences of gates in 5 different environments. 
Each frame is annotated with the corners of all gates visible in the image using the open source image annotation software labelme\footnote{https://github.com/wkentaro/labelme}, which is extended with KLT-Tracking for semi-automatic labelling.
The resulting dataset used for training consists of 28k images and is split into 24k samples for training and 4k samples for validation.
At training time, the data is augmented using random rotations of up to $\SI{30}{\degree}$ and random changes in brightness, hue, contrast and saturation.
}

\subsection{Network Architecture and Deployment}
\auro{The network architecture is designed to optimally trade-off between computation time and accuracy. 
By conducting a neural network architecture search, the best performing architecture for the task is identified.
The architecture search is limited to variants of U-Net~\cite{ronneberger2015u} due to its ability to perform segmentation tasks efficiently with a very limited amount of labeled training data. 
The best performing architecture is identified as a 5-level U-Net with $[12, 18, 24, 32, 32]$ convolutional filters of size $[3, 3, 3, 5, 7]$ per level and a final additional layer operating on the output of the U-Net containing 12 filters.}
At each layer, the input feature map is zero-padded to preserve a constant height and width throughout the network. 
As activation function, LeakyReLU with $\alpha = 0.01$ is used.
For deployment on the Jetson Xavier, the network is ported to TensorRT~5.0.2.6. 
To optimize memory footprint and inference time, inference is performed in half-precision mode (FP16) and batches of two images \revision{of size $592\times352$} are fed to the network. 

\section{State Estimation}\label{sec:state_estimation}
The nonlinear measurement models of the VIO, gate detection, and laser rangefinder are fused using an EKF \cite{kalman1960new}.
In order to obtain the best possible pose accuracy relative to the gates, the EKF estimates the translational and rotational misalignment of the VIO origin frame, $\V$, with respect to the inertial frame, $\I$, represented by $\bm{p}_{V}$ and $\bm{q}_{\I\V}$, jointly with the gate positions, $\bm{p}_{G_i}$, and gate heading, $\varphi_{\I\G_i}$.
It can thus correct for an imprecise initial position estimate, VIO drift, and uncertainty in gate positions.
The EKF's state space at time $t_k$ is $\bm{x}_k = \bm{x}(t_k)$ with covariance $\bm{P}_k$ described by
\begin{align}
\bm{x}_k = \left(
    \bm{p}_{\V},
    \bm{q}_{\I\V},
    \bm{p}_{G_0},
    \varphi_{\I\G_0},
    \dots,
    \bm{p}_{G_{N-1}},
    \varphi_{\I\G_{N-1}}\right).\label{eqn:state_space}
\end{align}
The drone's corrected pose, $\left(\bm{p}_B, \bm{q}_{\I\B}\right)$, can then be computed from the VIO estimate, $\left(\bm{p}_{V\!B}, \bm{q}_{\V\B}\right)$, by transforming it from frame $\V$ into the frame $\I$ using $\left(\bm{p}_{V}, \bm{q}_{\I\V}\right)$ as
\auro{\begin{align}
    \bm{p}_\B &= \bm{p}_\V + \bm{R}_{\I\V} \cdot \bm{p}_{\V\B}, &
    \bm{q}_{\I\B} &= \bm{q}_{\I\V} \cdot \bm{q}_{\V\B}.
\end{align}}

All estimated parameters are expected to be time-invariant but subject to noise and drift. This is modelled by a Gaussian random walk, simplifying the EKF process update to:
\begin{align}
\bm{x}_{k+1} &= \bm{x}_{k}, &
\bm{P}_{k+1} &= \bm{P}_{k} + \Delta t_k \bm{Q},
\end{align}
where $\bm{Q}$ is the random walk process noise.
For each measurement $\bm{z}_k$ with noise $\bm{R}$ the predicted \textit{a priori} estimate, $\bm{x}_k^-$, is corrected with measurement function, $\bm{h}(\bm{x}_k^-)$, and Kalman gain, $\bm{K}_k$, resulting in the \textit{a posteriori} estimate, $\bm{x}_k^+$, as
\begin{align}
\bm{K}_k &= \bm{P}_k^- \bm{H}_k^\intercal \left( \bm{H}_k \bm{P}_k^- \bm{H}_k^\intercal + \bm{R} \right)^{-1}, \nonumber \\
\bm{x}_k^+ &= \bm{x}_k^- + \bm{K}_k \left( \bm{z}_k - \bm{h}(\bm{x}_k^-) \right), \\
\bm{P}_k^+ &= \left( \bm{I} - \bm{K}_k \bm{H}_k \right) \bm{P}_k^-, \nonumber
\end{align}
with $\bm{h}(\bm{x}_k^-)$, the measurement function with Jacobian $\bm{H}_k$.~

\auro{However, the filter state includes a rotation quaternion constrained to unit norm, $\|\bm{q}_{\I\V}\| \overset{!}{=} 1$.
This is effectively an over-parameterization in the filter state space and can lead to poor linearization as well as underestimation of the covariance.}
To apply the EKFs linear update step on the over-parameterized quaternion, it is lifted to its tangent space description, similar to \cite{Forster17troOnmanifold}.
The quaternion $\bm{q}_{\I\V}$ is composed of a reference quaternion, $\bm{q}_{\I\V_\des}$, which is adjusted after each update step, and an error quaternion, $\bm{q}_{\V_\des \V}$, of which only its vector part, $\tilde{\bm{q}}_{\V_\des \V}$, is in the EKF's state space.
\auro{Therefore we get
\begin{align}
    \bm{q}_{\I\V} &= \bm{q}_{\I\V_\des} \cdot \bm{q}_{\V_\des \V} &
    \bm{q}_{\V_\des \V} &=
        \begin{bmatrix}
        \sqrt{ 1 - \tilde{\bm{q}}^\intercal_{\V_\des \V} \cdot \tilde{\bm{q}}_{\V_\des \V}} \\
        \tilde{\bm{q}}_{\V_\des \V}
        \end{bmatrix}
        \label{eqn:quaternion_error}
\end{align}
from which we can derive the Jacobian of any measurement function, $\bm{h}(\bm{x})$, with respect to $\bm{q}_{\I\V}$ by the chain rule as
\begin{align}
    \part{}{\tilde{\bm{q}}_{\V_\des \V}} \bm{h}(\bm{x})
    &=
    \part{}{\bm{q}_{\I\V}} \bm{h}(\bm{x}) \cdot
    \part{\bm{q}_{\I\V}}{\tilde{\bm{q}}_{\V_\des \V}}
    \label{eqn:quaternion_chainrule} \\
    &=
    \part{}{\tilde{\bm{q}}_{\I\V}} \bm{h}(\bm{x}) \cdot
    [\bm{q}_{\I\V_\des}]_\times
    \begin{bmatrix}
        \frac{-\tilde{\bm{q}}^\intercal_{\V_\des \V}}{\sqrt{ 1 - \tilde{\bm{q}}^\intercal_{\V_\des \V} \cdot \tilde{\bm{q}}_{\V_\des \V}}} \\ 
    \bm{I}^{3 \times 3}
    \end{bmatrix}
    \label{eqn:quaternion_jacobian}
\end{align}
where we arrive at \eqref{eqn:quaternion_jacobian} by using \eqref{eqn:quaternion_error} in \eqref{eqn:quaternion_chainrule} and use $[\bm{q}_{\I\V_\des}]_\times$ to represent the matrix resulting from a lefthand-side multiplication with $\bm{q}_{\I\V_\des}$.
}

\subsection{Measurement Modalities}
\auro{All measurements up to the camera frame time $t_k$ are passed to the EKF together with the VIO estimate, $\bm{p}_{V\!B,k}$ and $\bm{q}_{\V\B,k}$, with respect to the VIO frame $\V$.
Note thate the VIO estimate is assumed to be a constant parameter, not a filter state, which vastly simplifies derivations ad computation, leading to an efficient yet robust filter.}

\subsubsection{Gate Measurements}
Gate measurements consist of the image pixel coordinates, $\bm{s}_{Co_{ij}}$, of a specific gate corner.
These corners are denoted with top left and right, and bottom left and right, as in ${j \in \mathcal{C}_j}$, ${\mathcal{C}_j := \lbrace _{TL}, _{TR}, _{BL}, _{BR} \rbrace}$ and the gates are enumerated by $i \in [0, N-1]$.
All gates are of equal width, $w$, and height, $h$, so that the corner positions in the gate frame, $\G_i$, can be written as $\bm{p}_{{G_i}Co_{ij}} = \frac{1}{2}\left(0, \pm w, \pm h\right)$.
The measurement equation can be written as the pinhole camera projection~\cite{Szeliski10book} of the gate corner into the camera frame.
A pinhole camera maps the gate corner point, $\bm{p}_{Co_{ij}}$, expressed in the camera frame, $\C$, into pixel coordinates as
\begin{align}
\bm{h}_\mathrm{Gate}(\bm{x}) &= \bm{s}_{Co_{ij}} = \frac{1}{\lbrack\bm{p}_{Co_{ij}}\rbrack_z}
\begin{bmatrix}
f_x & 0 & c_x\\
0 & f_y & c_y
\end{bmatrix}
\bm{p}_{Co_{ij}},
\label{eqn:projection}
\end{align}
where $[\cdot]_z$ indicates the scalar $z$-component of a vector, $f_x$ and $f_y$ are the camera's focal lengths and $\left(c_x,c_y\right)$ is the camera's optical center.
The gate corner point, $\bm{p}_{Co_{ij}}$, is given by
\begin{align}
\bm{p}_{Co_{ij}} =& \bm{R}_{\I\C}^\intercal \left( \bm{p}_{G_i} + \bm{R}_{\I \G_i} \bm{p}_{{G_i}Co_j} - \bm{p}_{C} \right),   
\end{align}
with $\bm{p}_C$ and $\bm{R}_{\I\C}$ being the transformation between the inertial frame $\I$ and camera frame $\C$,
\begin{align}
\bm{p}_{C} =&
    \bm{p}_{V} + \bm{R}_{\I\V} \left(
    \bm{p}_{V\!B} + \bm{R}_{\V\B} \bm{p}_{BC} \right),\\
\bm{R}_{\I\C} =& \bm{R}_{\I\V} \bm{R}_{\V\B} \bm{R}_{\B\C},
\end{align}
where $\bm{p}_{BC}$ and $\bm{R}_{\B\C}$ describe a constant transformation between the drone's body frame $\B$ and camera frame $\C$ (see Fig.~\ref{fig:drone_dynamics}).
The Jacobian with respect to the EKF's state space is derived using the chain rule,
\begin{align}
\part{}{\bm{x}} \bm{h}_\mathrm{Gate}(\bm{x}) &=
    \part{\bm{h}_\mathrm{Gate}(\bm{x})}{\bm{p}_{Co_{ij}}(\bm{x})}
    \cdot \part{\bm{p}_{Co_{ij}}(\bm{x})}{\bm{x}},
\end{align}
\auro{where the first term representing the derivative of the projection, and the second term represents the derivative with respect to the state space including gate position and orientation, and the frame alignment, which can be further decomposed using \eqref{eqn:quaternion_chainrule}.}

\subsubsection{Gate Correspondences}
\auro{
The gate detection (see Figure \ref{sec:gate_detection}) provides sets of $m$ measurements, $\bm{S}_{\hat{i}} = \{ \bm{s}_{Co_{\hat{i}j} 0}, \dots, \bm{s}_{Co_{\hat{i}j} m-1} \}$, corresponding to the unknown gate $\hat{i}$ at known corners $j \in \mathcal{C}_j$.
To identify the correspondences between a detection set $\bm{S}_{\hat{i}}$ and the gate $G_i$ in our map, we use the square sum of reprojection error.
For this, we first compute the reprojection of all gate corners, $\bm{s}_{Co_{ij}}$, according to \eqref{eqn:projection} and then compute the square error sum between the measurement set, $\bm{S}_{\hat{i}}$, and the candidates, $\bm{s}_{Co_{ij}}$.
Finally, the correspondence is established to the gate $G_i$ which minimizes the square error sum, as in
\begin{align}
    \underset{i \in [0, N-1]}{argmin}
    \sum_{ \bm{s}_{Co_{\hat{i}j}} \in \bm{S}_{\hat{i}}} 
    (\bm{s}_{Co_{\hat{i}j}} - \bm{s}_{Co_{ij}})^\intercal (\bm{s}_{Co_{\hat{i}j}} - \bm{s}_{Co_{ij}})
\end{align}
}

\subsubsection{Laser Rangefinder Measurement}
The drone's laser rangefinder measures the distance along the drones negative $z$-axis to the ground, which is assumed to be flat and at a height of $\SI{0}{\meter}$.
The measurement equation can be described by
\begin{align}
h_\mathrm{LRF}(\bm{x}) &= \frac{ [ \bm{p}_{B} ]_z }{ [ \bm{R}_{\I\B} \bm{e}_z^\B ]_z }
    = \frac{ [ \bm{p}_{V} + \bm{R}_{\I\V} \bm{p}_{VB} ]_z }{ [ \bm{R}_{\I\V} \bm{R}_{\V\B} \bm{e}_z^\B ]_z }.
\end{align}
\auro{The Jacobian with respect to the state space is again derived by $\part{h_\mathrm{LRF}}{\bm{p}_{V}}$ and $\part{h_\mathrm{LRF}}{\bm{q}_{\I\V}}$ and further simplified using \eqref{eqn:quaternion_chainrule}.}

\section{Path Planning}\label{sec:path_planning}
\auro{For the purpose of path planning, the drone is assumed to be a point mass with bounded accelerations as inputs.
This simplification allows for the computation of time-optimal motion primitives in closed-form and enables the planning of approximate time-optimal paths through the race course in real time.
Even though the dynamics of the quadrotor vehicle's acceleration cannot be neglected in practice, it is assumed that this simplification still captures the most relevant dynamics for path planning and that the resulting paths approximate the true time-optimal paths well.
In order to facilitate the tracking of the approximate time-optimal path, polynomials of order four are fitted to the path which yield smoother position, velocity and acceleration commands, and can therefore be better tracked by the drone.

In the following, time-optimal motion primitives based on the simplified dynamics are first introduced and then a path planning strategy based on these motion primitives is presented.
Finally, a method to parameterize the time-optimal path is introduced.}

\subsection{Time-Optimal Motion Primitive}\label{subsec:motion_primitive}
The minimum times, $T_x^*$, $T_y^*$ and $T_z^*$, required for the vehicle to fly from an initial state, consisting of position and velocity, to a final state while satisfying the simplified dynamics ${\ddot{\bm{p}}_B(t) = \bm{u}(t)}$ with the input acceleration $\bm{u}(t)$ being constrained to $\underline{\bm{u}} \leq \bm{u}(t) \leq \overline{\bm{u}}$ are computed for each axis individually.
Without loss of generality, only the $x$-axis is considered in the following.
Using Pontryagin's maximum principle~\cite{bertsekas1995dynamicprogramming}, it can be shown that the optimal control input is bang-bang in acceleration, i.e., has the form
\begin{align}
u_x^*(t) = \begin{cases}
\underline{u}_x, & 0 \leq t \leq t^*,\\
\overline{u}_x, & t^* < t \leq T_x^*,
\end{cases}
\label{eq:mp_sol_unconstr}
\end{align}
or vice versa with the control input first being $\overline{u}_x$ followed by $\underline{u}_x$. 
In order to control the maximum velocity of the vehicle, e.g., to constrain the solutions to ranges where the simplified dynamics approximate the true dynamics well or to limit the motion blur of the camera images, a velocity constraint of the form $\underline{\bm{v}}_B \leq \bm{v}_B(t) \leq \overline{\bm{v}}_B$ can be added, in which case the optimal control input has bang-singular-bang solution~\cite{maurer1977optimalcontrol}
\begin{align}
u_x^*(t) = \begin{cases}
\underline u_x, & 0 \leq t \leq t_1^*,\\
0, & t_1^* < t \leq t_2^*,\\
\overline u_x, & t_2^* < t \leq T_x^*,\\
\end{cases}
\label{eq:mp_sol_constr}
\end{align}
or vice versa.
It is straightforward to verify that there exist closed-form solutions for the minimum time, $T_x^*$, as well as the switching times, $t^*$, in both cases \eqref{eq:mp_sol_unconstr} or \eqref{eq:mp_sol_constr}.

\begin{figure}[t]
	\centering
	\includegraphics[width=\columnwidth]{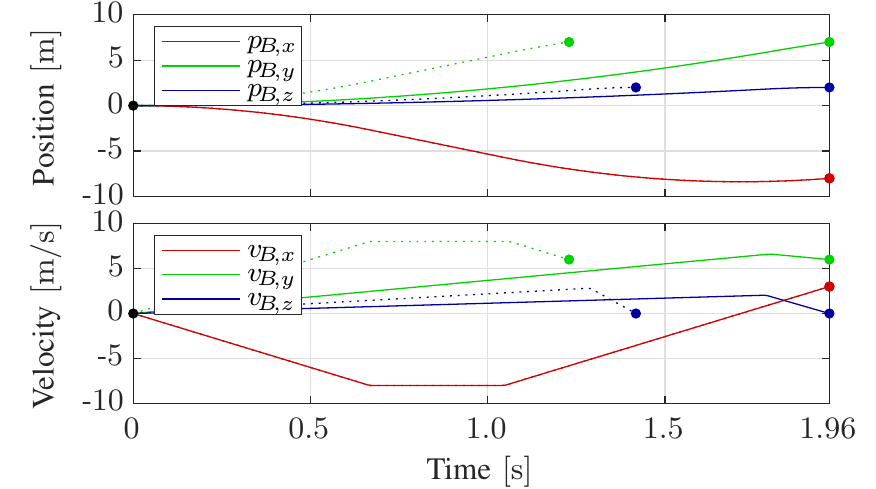}
	\caption{Example time-optimal motion primitive starting from rest at the origin to a random final position with non-zero final velocity. The velocities are constrained to $\pm 7.5\si[per-mode=symbol]{\meter\per\second}$ and the inputs to $\pm 12\si[per-mode=symbol]{\meter\per\second^2}$. The dotted lines denote the per-axis time-optimal maneuvers.}
	\label{fig:mp_sol}
\end{figure}

Once the minimum time along each axis is computed, the maximum minimum time, $T^* = \mathrm{max}(T_x^*,T_y^*,T_z^*)$, is computed and motion primitives of the same form as in \eqref{eq:mp_sol_unconstr} or \eqref{eq:mp_sol_constr} are computed among the two faster axes but with the final time constrained to $T^*$ such that trajectories along each axis end at the same time. In order for such a motion primitive to exist, a new parameter $\alpha \in \lbrack 0, 1\rbrack$ is introduced that scales the acceleration bounds, i.e., the applied control inputs are scaled to $\alpha \underline{u}_x$ and $\alpha \overline{u}_x$, respectively. 
Fig.~\ref{fig:mp_sol} depicts the position and velocity of an example time-optimal motion primitive.

\subsection{Sampling-Based Receding Horizon Path Planning}
The objective of the path planner is to find the time-optimal path from the drone's current state to the final gate, passing through all the gates in the correct order.
Since the previously introduced motion primitive allows the generation of time-optimal motions between any initial and any final state, the time-optimal path can be planned by concatenating a time-optimal motion primitive starting from the drone's current (simplified) state to the first gate with time-optimal motion primitives that connect the gates in the correct order until the final gate.
This reduces the path planning problem to finding the drone's optimal state at each gate such that the total time is minimized.
To find the optimal path, a sampling-based strategy is employed where states at each gate are randomly sampled and the total time is evaluated subsequently.
In particular, the position of each sampled state at a specific gate is fixed to the center of the gate and the velocity is sampled uniformly at random such the velocity lies within the constraints of the motion primitives and the angle between the velocity and the gate normal does not exceed a maximum angle, $\varphi_\mathrm{max}$
It is trivial to show that as the number of sampled states approaches infinity, the computed path converges to the time-optimal path.

In order to solve the problem efficiently, the path planning problem is interpreted as a shortest path problem.
At each gate, $M$ different velocities are sampled and the arc length from each sampled state at the previous gate is set to be equal to the duration, $T^*$, of the time-optimal motion primitive that guides the drone from one state to the other.
Due to the existence of a closed-form expression for the minimum time, $T^*$, setting up and solving the shortest path problem can be done very efficiently using, e.g., Dijkstra's algorithm~\cite{bertsekas1995dynamicprogramming} \auro{resulting in the optimal path $\bm{p}^*(t)$}.
In order to further reduce the computational cost, the path is planned in a receding horizon fashion, i.e., the path is only planned through the next $N$ gates.

\auro{
\subsection{Path Parameterization}
Due to the simplifications of the dynamics that were made when computing the motion primitives, the resulting path is infeasible with respect to the quadrotor dynamics \eqref{eq:trans_dynamics} and \eqref{eq:att_kinematics} and thus is impossible to be tracked accurately by the drone.
To simplify the tracking of the time-optimal path, the path is approximated by fourth order polynomials in time.
In particular, the path is divided into multiple segments of equal arc length.
Let $t\in\lbrack t_{k}, t_{k+1})$ be the time interval of the $k$-th segment.
In order to fit the polynomials, $\bar{\bm{p}}_k(t)$, to the $k$-th segment of the time-optimal path, we require that the initial and final position and velocity are equal to those of the time-optimal path, i.e.,
\begin{align}
    \bar{\bm{p}}_k(t_k) &= \bm{p}^*(t_k),&
    \bar{\bm{p}}_k(t_{k+1}) &= \bm{p}^*(t_{k+1}),\\
    \dot{\bar{\bm{p}}}_k(t_k) &= \dot{\bm{p}}^*(t_k),&
    \dot{\bar{\bm{p}}}_k(t_{k+1}) &= \dot{\bm{p}}^*(t_{k+1}),
\end{align}
and that the positions at $t=\left(t_{k+1} - t_{k}\right)/2$ coincide as well:
\begin{align}
    \bar{\bm{p}}_k\left(\frac{t_{k+1} + t_{k}}{2}\right) &= \bm{p}^*\left(\frac{t_{k+1} + t_{k}}{2}\right).
\end{align}
The polynomial parameterization $\bar{\bm{p}}_k(t)$ of the $k$-th segment is then given by
\begin{align}
    \bar{\bm{p}}_k(t) = \bm{a}_{4,k} s^4 + \bm{a}_{3,k} s^3 + \bm{a}_{2,k} s^2 + \bm{a}_{1,k} s + \bm{a}_{0,k},
    \label{eq:polynomial_param}
\end{align}
with $s = t - t_k$ being the relative time since the start of $k$-th segment. The velocity and acceleration required for the drone to track this polynomial path can be computed by taking the derivatives of \eqref{eq:polynomial_param}, yielding
\begin{align}
    \dot{\bar{\bm{p}}}_k(t) &= 4 \bm{a}_{4,k} s^3 + 3 \bm{a}_{3,k} s^2 + 2 \bm{a}_{2,k} s + \bm{a}_{1,k},\\
    \ddot{\bar{\bm{p}}}_k(t) &= 12 \bm{a}_{4,k} s^2 + 6 \bm{a}_{3,k} s + 2 \bm{a}_{2,k}.
\end{align}
}

\section{Control}\label{sec:control}
\begin{figure*}[t]
	\centering
	\includegraphics[width=\textwidth]{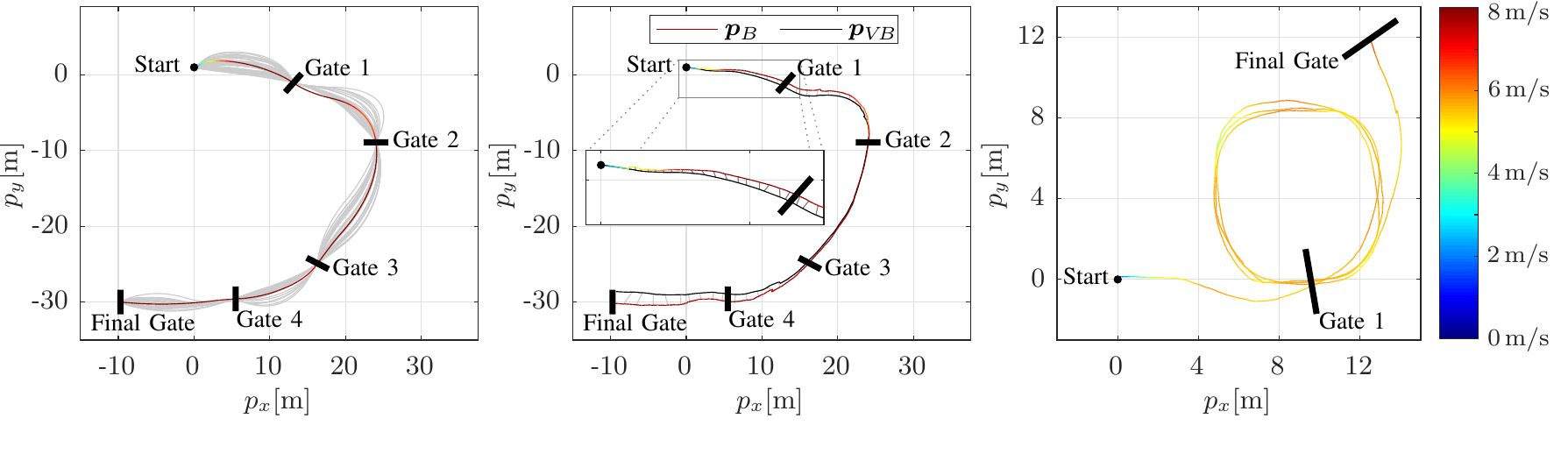}
	\caption{Top view of the planned (left) and executed (center) path at the championship race, and an executed multi-lap path at a testing facility (right).
	Left: Fastest planned path in color, sub-optimal sampled paths in gray.
	Center: VIO trajectory as $\bm{p}_{V\!B}$ and corrected estimate as $\bm{p}_B$.}
	\label{fig:results}
\end{figure*}
This section presents a control strategy to track the near time-optimal path from Section \ref{sec:path_planning}.
The control strategy is based on a cascaded control scheme with an outer position control loop and an inner attitude control loop, where the position control loop is designed under the assumption that the attitude control loop can track setpoint changes perfectly, i.e., without any dynamics or delay.

\subsection{Position Control}
The position control loop along the inertial $z$-axis is designed such that it responds to position errors
$$
{p_{B_\err,z} = p_{B_\des,z} - p_{B,z}}
$$
in the fashion of a second-order system with time constant $\tau_{\pos,z}$ and damping ratio $\zeta_{\pos,z}$,
\begin{align}
\ddot{p}_{B,z} &= \frac{1}{\tau_{\pos,z}^2}p_{B_\err,z} +  \frac{2\zeta_{\pos,z}}{\tau_{\pos,z}}\dot{p}_{B_\err,z} + \ddot{p}_{B_\des,z}.
\label{eq:des_position_dynamics}
\end{align}
Similarly, two control loops along the inertial $x$- and $y$-axis are shaped to make the horizontal position errors behave like second-order systems with time constants ${\tau_{\pos,xy}}$ and damping ratio ${\zeta_{\pos,xy}}$.
Inserting~\eqref{eq:des_position_dynamics} into the translational dynamics~\eqref{eq:trans_dynamics}, the total thrust, $f$, is computed to be
\begin{align}
f &= \frac{\lbrack m\left(\ddot{\bm{p}}_{B_\des} + \bm{g}\right) + \bm{R}_{\I\B}\bm{D}\bm{R}_{\I\B}^\intercal \bm{v}_{B}\rbrack_z}{\lbrack R_{\I\B \bm{e}_z^\B}\rbrack_z}.
\end{align}

\subsection{Attitude Control}
The required acceleration from the position controller determines the orientation of the drone's $z$-axis and is used, in combination with a reference yaw angle, $\varphi_\des$, to compute the drone's reference attitude.
The reference yaw angle is chosen such that the drone's $x$-axis points towards the reference position \SI{5}{\meter} ahead of the current position, i.e., that the drone looks in the direction it flies.
A nonlinear attitude controller similar to~\cite{brescianini2018attitude} is applied that prioritizes the alignment of the drone's $z$-axis, which is crucial for its translational dynamics, over the correction of the yaw orientation:
\begin{align}
\bm{\omega} = \frac{2~\mathrm{sgn}(q_w)}{\sqrt{q_w^2 + q_z^2}}
\bm{\mathrm{T}}_\att^{-1}
\begin{bmatrix}
q_w q_x - q_y q_z\\
q_w q_y + q_x q_z\\
q_z
\end{bmatrix},
\end{align}
where $q_w$, $q_x$, $q_y$ and $q_z$ are the components of the attitude error, $\bm{q}_{\I\B}^{-1}\otimes\bm{q}_{\I\B_\des}$, and where $\bm{\mathrm{T}}_\att$ is a diagonal matrix containing the per-axis first-order system time constants for small attitude errors.

\section{Results}\label{sec:results}

The proposed system was used to race in the \textit{2019 AlphaPilot} championship race. The course at the championship race consisted of five gates and had a total length of \SI{74}{\meter}.
A top view of the race course as well as the results of the path planning and the fastest actual flight are depicted in Fig.~\ref{fig:results} (left and center).
With the motion primitive's maximum velocity set to $\SI[per-mode=symbol]{8}{\meter\per\second}$, the drone successfully completed the race course in a total time of $\SI{11.36}{\second}$, with only two other teams also completing the full race course.
The drone flew at an average velocity of $\SI[per-mode=symbol]{6.5}{\meter\per\second}$ and reached the peak velocity of $\SI[per-mode=symbol]{8}{\meter\per\second}$ multiple times.
Note that due to missing ground truth, Fig. \ref{fig:results} only shows the estimated and corrected drone position.

The system was further evaluated at a testing facility where there was sufficient space for the drone to fly multiple laps (see Fig.~\ref{fig:results}, right), albeit the course consisted of only two gates.
The drone was commanded to pass four times through \textit{gate 1} before finishing in the \textit{final gate}.
Although the gates were not visible to the drone for most of the time, the drone successfully managed to fly multiple laps.
Thanks to the global gate map and the VIO state estimate, the system was able to plan and execute paths to gates that are not directly visible.
By repeatedly seeing either one of the two gates, the drone was able to compensate for the drift of the VIO state estimate, allowing the drone to pass the gates every time exactly through their center.
Note that although seeing \textit{gate 1} in Fig.~\ref{fig:results} (right) at least once was important in order to update the position of the gate in the global map, the VIO drift was also estimated by seeing the \textit{final gate}.

The results of the system's main components are discussed in detail in the following subsections, and a video of the results is attached to the paper.

\subsection{Gate Detection}
\label{sec:res_gatedetection}

\auro{\textbf{Architecture Search:} Due to the limited computational budget of the Jetson Xavier, the network architecture was designed to maximize detection accuracy while keeping a low inference time. To find such architecture, different variants of U-Net~\cite{ronneberger2015u} are compared. 
Table~\ref{tab:nw_architectures} summarizes the performance of different network architectures.
Performance is evaluated quantitatively on a separate test set of 4k images with respect to intersection over union (IoU) and precision/recall for corner predictions.
While the IoU score only takes full gate detections into account, the precision/recall scores are computed for each corner detection. 
Based on these results, architecture UNet-5 is selected for deployment on the real drone due to the low inference time and high performance.
On the test set, this network achieves an IoU score with the human-annotated ground truth of $96.4$\%. When only analyzing the predicted corners, the network obtains a precision of $0.997$ and a recall of $0.918$.

\begin{table}[t]
\centering
\caption{Comparison of different network architectures with respect to intersection over union (IoU), precision (Pre.) and recall (Rec.). The index in the architecture name denotes the number of levels in the U-Net. All networks contain one layer per level with kernel sizes of $[3, 3, 5, 7, 7]$ and $[ 12, 18, 24, 32, 32]$ filters per level. Architectures labelled with 'L' contain twice the amount of filters per level. Timings are measured for single input images of size 352x592 on a desktop computer equipped with an NVIDIA RTX 2080 Ti.}
\label{tab:nw_architectures}
\begin{tabular}{lccccc}
\toprule
\textbf{Arch.} &
  \textbf{IoU} &
  \textbf{Pre.} &
  \textbf{Rec.} &
  \textbf{\#params} &
  \textbf{latency [s]} \\ \midrule
UNet-5L & 0.966 & 0.997 & 0.967 & 613k &  0.106\\
UNet-5 & 0.964 & 0.997 & 0.918 & 160k &  0.006\\
UNet-4L & 0.948 & 0.997 & 0.920 & 207k & 0.085  \\
UNet-4 & 0.941 & 0.989 & 0.862 & 58k & 0.005\\
UNet-3L &  0.913    &   0.991   &   0.634   &  82k & 0.072 \\
UNet-3 & 0.905 & 0.988 & 0.520 & 27k & 0.005\\
\bottomrule
\end{tabular}
\end{table}

\noindent \textbf{Deployment:}}
Even in instances of strong changes in illumination, the gate detector was able to accurately identify the gates in a range of $2-\SI{17}{\meter}$. Fig.~\ref{fig:network_prediction} illustrates the quality of detections during the championship race (1st row) as well as for cases with multiple gates, represented in the test set (2nd/3rd row).
With the network architecture explained in Section~\ref{sec:gate_detection}, one simultaneous inference for the left- and right-facing camera requires computing  \SI{3.86}{\giga FLOPS} (\SI{40}{\kilo FLOPS} per pixel). 
By implementing the network in TensorRT and performing inference in half-precision mode (FP16), this computation takes \SI{10.5}{\milli\second} on the Jetson Xavier and can therefore be performed at the camera update rate.

\subsection{State Estimation}
Compared to a pure VIO-based solution, the EKF has proven to significantly improve the accuracy of the state estimation relative to the gates.
As opposed to the works by \cite{Li19arxiv,Jun18ral,Kaufmann18icra}, the proposed EKF is not constrained to only use the next gate, but can work with any gate detection and even profits from multiple detections in one image.
Fig.~\ref{fig:results} (center) depicts the flown trajectory estimated by the VIO system as~$\bm{p}_{V\!B}$ and the EKF-corrected trajectory as~$\bm{p}_{B}$ (the estimated corrections are depicted in gray).
Accumulated drift clearly leads to a large discrepancy between VIO estimate~$\bm{p}_{V\!B}$ and the corrected estimate~$\bm{p}_{B}$.
Towards the end of the track at the two last gates this discrepancy would be large enough to cause the drone to crash into the gate.
However, the filter corrects this discrepancy accurately and provides a precise pose estimate relative to the gates.
Additionally, the imperfect initial pose, in particular the yaw orientation, is corrected by the EKF while flying towards the first gate as visible in the zoomed section in Fig.~\ref{fig:results} (center).

\subsection{Planning and Control}
\label{sec:res_planning}

\begin{table}[t]
    \centering
    \caption{Total flight time vs. computation time averaged over 100 runs. The percentage in parenthesis is the computation time with respect to the computational time for the full track.}
    \label{tab:receding_horizon}
    \begin{tabular}{lcc}
        \toprule
        $\bm{N_{gates}}$ & \textbf{flight time} & \textbf{computation time} \\
        \midrule
        $1$ & $\SI{9.5935}{\second}$ & $\SI{1.66}{\milli\second}$ $2.35\%$ \\
        \midrule
        $2$ & $\SI{9.2913}{\second}$ & $\SI{18.81}{\milli\second}$ $26.56\%$ \\
        \midrule
        $3$ & $\SI{9.2709}{\second}$ & $\SI{35.74}{\milli\second}$ $50.47\%$ \\
        \midrule
        $4$ & $\SI{9.2667}{\second}$ & $\SI{53.00}{\milli\second}$ $74.84\%$ \\
        \midrule
         $5$ (full track) & $\SI{9.2622}{\second}$ & $\SI{70.81}{\milli\second}$ $100\%$ \\
        \midrule
         CPC \cite{Foehn2020cpc} (full track) & $\SI{6.520}{\second}$ & $4.62 \cdot 10^5\si{\milli\second}$ $6524\%$ \\
        \bottomrule
    \end{tabular}
\end{table}

Fig.~\ref{fig:results} (left) shows the nominally planned path for the \textit{AlphaPilot} championship race, where the coloured line depicts the fastest path along all the sampled paths depicted in gray.
In particular, a total of $M = 150$ different states are sampled at each gate, with the velocity limited to $\SI[per-mode=symbol]{8}{\meter\per\second}$ and the angle between the velocity and the gate normal limited to $\varphi_\mathrm{max}=\SI{30}{\degree}$.
During flight, the path is re-planned in a receding horizon fashion through the next $N=3$ gates (see Fig.~\ref{fig:results}, center).
It was experimentally found that choosing $N \geq 3$ only has minimal impact of the flight time comapred to planning over all gates,  while greatly reducing the computational cost.
\auro{
Table \ref{tab:receding_horizon} presents the trade-offs between total flight time and computation cost for different horizon lengths $N$ for the track shown in Fig.~\ref{fig:results} (left).
\revision{In addition, Table \ref{tab:receding_horizon} shows the flight and computation time of the time-optimal trajectory generation from \cite{Foehn2020cpc}, which significantly outperforms our approach but is far away from real-time execution with a computation time of $\SI{462}{\second}$ for a single solution.
Online replanning would therefore not be possible, and any deviations from the nominal track layout could lead to a crash.}

Please also note that the evaluation of our method is performed in Matlab on a laptop computer, while the final optimized implementation over $N=3$ gates achieved replanning times of less than $\SI{2}{\milli\second}$ on the Jetson Xavier and can thus be done in every control update step.
}
Fig.~\ref{fig:results} (right) shows resulting path and velocity of the drone in a multi-lap scenario, where the drone's velocity was limited to $\SI[per-mode=symbol]{6}{\meter\per\second}$. It can be seen that drone's velocity is decreased when it has to fly a tight turn due to its limited thrust.
\section{Discussion and Conclusion}\label{sec:conclusion}
The proposed system managed to complete the course at a velocity of $\SI[per-mode=symbol]{5}{\meter\per\second}$ with a success rate of $100\%$ and at $\SI[per-mode=symbol]{8}{\meter\per\second}$ with a success rate of $60\%$.
At higher speeds, the combination of VIO tracking failures and no visible gates caused the drone to crash after passing the first few gates.
This failure could be caught by integrating the gate measurements directly in a VIO pipeline, tightly coupling all sensor data.
Another solution could be a perception-aware path planner trading off time-optimality against motion blur and maximum gate visibility.

The advantages of the proposed system are (\rom{1}) a drift-free state estimate at high speeds, (\rom{2}) a global and consistent gate map, and (\rom{3}) a real-time capable near time-optimal path planner.
However, these advantages could only partially be exploited as the races neither included multiple laps, nor had complex segments where the next gates were not directly visible.
Nevertheless, the system has proven that it can handle these situations and is able to navigate through complex race courses reaching speeds up to $\SI[per-mode=symbol]{8}{\meter\per\second}$ and completing the championship race track of $\SI{74}{\meter}$ in $\SI{11.36}{\second}$.

While the \textit{2019 AlphaPilot Challenge} pushed the field of autonomous drone racing, in particularly in terms of speed, autonomous drones are still far away from beating human pilots.
Moreover, the challenge also left open a number of problems, most importantly that the race environment was partially known and static without competing drones or moving gates.
In order for autonomous drones to fly at high speeds outside of controlled or known environments and succeed in many more real-world applications, they must be able to handle unknown environments, perceive obstacles and react accordingly.
These features are areas of active research and are intended to be included in future versions of the proposed drone racing system.

\balance
\bibliographystyle{unsrtnat}
\bibliography{references}

\end{document}